\documentclass[10pt,twocolumn,letterpaper]{article}

\usepackage{3dv}
\usepackage{times}
\usepackage{epsfig}
\usepackage{graphicx}
\usepackage{amsmath}
\usepackage{amssymb}

\newcommand{\template}{\mathcal{T}}
\newcommand{\templateSet}{\mathcal{S}}
\newcommand{\refImage}{O}
\newcommand{\testImage}{I}
\newcommand{\locationList}{\mathcal{R}}
\newcommand{\location}{r}
\newcommand{\shift}{c}
\newcommand{\similarity}{\mathcal{E}}
\newcommand{\feat}{\Psi}
\newcommand{\descriptor}{v}
\newcommand{\object}{obj}
\newcommand{\pose}{pose}

\newcommand{\modalityList}{M}
\newcommand{\kdtree}{\textit{k}-d tree~}

\newcommand{\function}{f}
\newcommand{\scoring}{s}
\newcommand{\scoringConst}{\alpha}

\newcommand{\splitfunc}{h} 
\newcommand{\tree}{\mathcal{P}}
\newcommand{\featValue}{x}
\newcommand{\splitParam}{\theta}
\newcommand{\featSelect}{\phi}
\newcommand{\splitPlane}{\psi}
\newcommand{\splitThresh}{\tau}
\newcommand{\nodeIndex}{n}
\newcommand{\dimension}{d}
\newcommand{\exemplar}{\descriptor_e}
\newcommand{\energyFunc}{E}
\newcommand{\entropy}{\mathcal{H}}
\newcommand{\informationGain}{\mathcal{I}}
\newcommand{\completeness}{\lambda}
\newcommand{\splitParamPool}{\Theta}
\newcommand{\objMask}{\gamma}
\newcommand{\featSpace}{\mathcal{V}}
\newcommand{\featCount}{D}
\newcommand{\isEqual}{\delta}
\newcommand{\fuzzyFactor}{\xi}

\newcommand{\rsplitfunc}{\check{\splitfunc}} 
\newcommand{\rsplitParam}{\check{\splitParam}} 
\newcommand{\rfeatSelect}{\check{\featSelect}} 
\newcommand{\isBackground}{\beta} 
\newcommand{\rsplitThresh}{\check{\tau}}
\newcommand{\rLUTThresh}{\rho}

\newcommand{\bigo}{\ensuremath{\mathcal{O}}}

\DeclareMathOperator*{\argmin}{argmin}
\DeclareMathOperator*{\argmax}{argmax}

\threedvfinalcopy 

\def\threedvPaperID{****} 

\ifthreedvfinal\fi
\begin{document}

\title{Real-time Background-aware 3D Textureless Object Pose Estimation}

\author{Mang Shao\\
Imperial College London\\
{\tt\small mang.shao08@imperial.ac.uk}
\and
Danhang Tang\\
Imperial College London\\
{\tt\small danhangtang@gmail.com}
\and
Tae-Kyun Kim\\
Imperial College London\\
{\tt\small tk.kim@imperial.ac.uk}
}

\maketitle

\begin{abstract}

In this work, we present a modified fuzzy decision forest for real-time 3D object pose estimation based on typical template representation. We employ an extra preemptive background rejector node in the decision forest framework to terminate the examination of background locations as early as possible, result in a significantly improvement on efficiency. Our approach is also scalable to large dataset since the tree structure naturally provides a logarithm time complexity to the number of objects. Finally we further reduce the validation stage with a fast breadth-first scheme. The results show that our approach outperform the state-of-the-arts on the efficiency while maintaining a comparable accuracy.
\end{abstract}

\section{Introduction}
%

In this paper, we focus on real-time 3D rigid object pose estimation with RGB-D data. Since rigid object pose has 6 degrees of freedom (x, y, z, pitch, yaw, roll), this problem is also known as 6-DoF pose estimation. A typical method of estimating 6-DoF object poses is through {\it template matching}, which requires a mesh model that usually obtained via scanning the target object. A large set of training `templates' with labels that uniformly covers the pose space is then generated by rendering the mesh model. During testing, a sufficiently similar template is found via a distance-based search process, often with approximate nearest neighbour (ANN) techniques. 

There are mainly three different ways of performing ANN on template matching: exhaustive, hashing-based and tree-based. Whilst given the feature descriptor, exhaustive method (\eg, LineMOD~\cite{Linemod}) guarantees to find the most similar match, its linear complexity is definitely not ideal. Although hashing-based methods have sublinear or even constant complexity during searching, the design of an efficient hash function with good trade-off between memory consumption and matching performance is not trivial. Tree-based methods to solve ANN problem, \eg, $k$-d tree, also significantly lowers the complexity. However, the cascade nature of tree structure makes them prone to be less robust to noise and the efficiency suffers from the curse of dimensionality due to {\it backtrack}.  

\section{Related work}
In this section we categorise existing 6-DoF pose estimation methods into distance-based, learning-based and registration-based, then we discuss good strategies of accelerating template matching procedure. 

\subsection{Distance-based}

Distance-based methods approach this problem by defining a distance metric to measure the similarity between samples. Then a set of samples from different view points, usually rendered from 3D object models, is generated as the training set. During testing, the object pose on a query image is retrieved by pairwise comparison between extracted templates and the training set. 

To effectively measure the similarity between object views, a compact and discriminative description vector is required. Hinterstoisser et al. \cite{Linemod_ACCV} present a novel image representation, a rigid template using colour gradient and surface normal as feature descriptors called LineMOD. The templates are synthetically rendered from 3D object mesh models under different scales and view angles. Similar to other traditional template matching approaches, each template matches with all possible locations across the query image to produce a similarity score map. Despite the exhaustive search, it achieves real-time speed for single object pose estimation.

Tree-based approaches apply binary search in multidimensional space. Given a query point and a set of data points, this approach partitions the search space into roughly halves in each iteration, until there is only one data point left in the search space. The complexity is therefore $O(\log n)$ to the number of data points. Though, k-d trees are generally considered not suitable for high-dimensional spaces searching as most of the points in the tree will be evaluated and the efficiency is no better than exhaustive search \cite{kdtree_book}. One improvement by Beis and Lowe \cite{kdtree_bbf}, called best-bin first algorithm, uses a backtracking strategy to prioritise the searching queue based on closeness and achieves two orders of magnitude speed up. Another solution applies randomness in building multiple trees to improve the search speed at the cost of the individual k-d tree not always returning the exact nearest neighbours.

Hash table is a well-known data structure that allows a symbol lookup in $O(1)$ complexity. In other words, the searching time is constant regardless of the database size. However, hash table is only able to find the exact match while in ANN searching problem we seek approximate matches. The most straight forward solution is hashing the whole quantised feature space into a single hash table so that every possible query point directly maps to their nearest data points. Unfortunately this naive approach is no longer feasible for high-dimensional data. A recent work Kehl et al. \cite{Hashmod} employs hashing techniques to achieve sublinear scalability by exploring different hashing key learning strategies and achieve sublinear complexity to the number of templates and outperform the state-of-the-art in terms of runtime.

\subsection{Learning-based} 
Learning-based methods usually generalise better to variations in viewpoint, translation and slight shape deformations.

The methods fall into this category focus on better generalisation to slight variations in translation, local shape and viewpoint. The explicit background/foreground separation is learnt parametrically to deal with heavy background clutter. The result shows these approaches cause less false positives than nearest neighbour approaches. However, the efficacy is their dependency on the quality of negative training samplesœ, and this benefit may not transfer across different domains. Tejani \etal~\cite{HoughForest} propose to incorporate a one-class learning scheme into the hough forest framework for 6-DoF problems. Rios-Cabrera and Tuytelaars \cite{DTT} extend LineMOD by learning the templates in a discriminative fashion and handle 10-30 3D objects at frame rates above 10fps using a single CPU core. 

\subsection{Registration-based} 

Registration-based methods attempt to fit a pose hypothesis to the observation, by iteratively update and minimise the discrepancy between the query sample and a sample rendered from the current pose hypothesis. A popular choice is the Iterative Closest Point (ICP)~\cite{ICP_fitz}. 

Johnson and Hebert present an early seminal work \cite{spin} for simultaneous recognition of multiple objects in scenes containing clutter and occlusion, based on matching surfaces by matching points using the spin image representation.
Gordon and Lowe \cite{lowe_2006} present a feature-based object pose estimation framework that accurately track camera using learned models and SIFT features \cite{SIFT}. The estimation is performed by matching query image features with 3D object model features and solving the Perspective-n-Point (PnP) problem for the 2D-to-3D correspondences. Drost et al. \cite{drost} propose a novel method that creates a global model description based on oriented point pair features and matches that model locally using a fast voting scheme. Another recent works \cite{Hao_2d-to-3d} improve the framework by introducing a novel matching scheme. 

\subsection{Accelerating Template Matching}

The first difficulty of accelerating template matching with efficient searching schemes comes from the high dimensionality. One single coordinate is far not representative enough to quickly reject candidates, thus leads to suboptimal performance. It is fortunate that many recent works have shown that with a well chosen feature descriptor, few coordinates have enough contrast to reliably find the match. LineMOD \cite{Linemod} achieves good performance by extracting only the best 100 dimensions out of ten thousand from each individual template to perform an optimised exhaustive search. However, it is not trivial to apply efficient searching algorithm based on this approach. Since the best 100 dimensions are in different subspaces so the distance measurement between them is not meaningful, and ten thousand dimensions are simply too large for most of ANN algorithms.

One feasible solution is to cluster the templates into few sets, which has been proposed in few recent works. Hashmod \cite{Hashmod} clusters the templates with randomised forest and employs hashing techniques; Discriminatively Trained Templates (DTT) \cite{DTT} clusters the templates with a bottom-up clustering method and constructs strong classifiers using AdaBoost. The underlying reason is the clustered subsets share common `relevant' feature dimensions, that is to say, the templates in a subset can be well classified using much less coordinates. 

Another difficulty is the heavy noises present due to the background clutter, occlusion and other environmental nuisances. Since the features in a template are generally local to a small region, it is very likely that the noises renders some feature dimensions completely irrelevant to the ground truth. It is necessary to exam multiple feature dimensions to increase the signal-to-noise-ratio, thus reliable matching results.

With the same reason, a final validation stage is inevitable to achieve a good precision-recall rate. In this stage, a full similarity measure is calculated between the testing image region and a small subset of templates. Compare to previous stages, validation is expensive and is usually the bottleneck of the whole method, due to much more feature dimensions are involved in the calculation. Therefore to reduce the computational cost, a good trade-off needs to be made between the size of validation subset and matching accuracy.

To sum up, a good approach to accelerate template matching should be: (i) reducing irrelevant feature dimensions; (ii) testing on multiple dimensions simultaneously to be less prone to outliers; (iii) reducing the validation subset as much as possible while maintain the matching robustness; and (iv) optimise validation process to further speed up. With these in mind, we propose our tree-based method to address the problems for efficient template matching.

\section{Method}

Decision tree is one of the most commonly known model in machine learning for sublinear nearest neighbour search. However, training a tree-based classifier using templates directly are problematic due to insufficient training samples and noisy feature dimensions. In recent works on 3D object pose estimation using tree-based classifier, both \cite{HoughForest,RF} use local feature-based representations instead of holistic template to alleviate the overfitting issue. However, this approach requires an additional geometric verification stage, Hough voting in \cite{HoughForest} and RANSAC-based optimisation in \cite{RF}, that likely to drag the process out too long to meet the requirement of real-time applications.

In this work, we propose several extensions to the classic random forest framework to significantly accelerate template matching with marginally loss in accuracy in comparison to the exhaustive search.

\subsection{Template Matching in Random Forest Framework}

Template matching approach describes each view of the object instance into a single template representation $\template$. The template is defined as $\template = (\refImage,\locationList)$, where $\refImage$ is a reference RGB-D image of an object and $\locationList$ denotes the set of locations $\location$ in $\refImage$. The similarity measure $\similarity$ between a template $\template$ and an input image $\testImage$ shifted by $\shift$ can be formalised as:
\begin{equation}
	\similarity(\testImage,\template,\shift) = \sum_{\location \in \locationList}{\|(\feat(\refImage,\location) - \feat(\testImage,\shift+\location))\|},
\end{equation}
where $\feat$ denotes the local feature descriptor, $\| \cdot \|$ denotes the distance function. Thus, the overall similarity is the sum of all individual corresponding local features differences.

Given an input RGB-D image, we use randomised decision forest $\{\tree\}$ to classify sliding windows centred at each pixel location $\shift$. Leaf node of each tree $\tree$ that the pixel ends up retrieves a set of template $\{\template\}$ so that a final classification is produced by a full validation.

\begin{figure}[h]
\centering
  \includegraphics[width=0.5\textwidth]{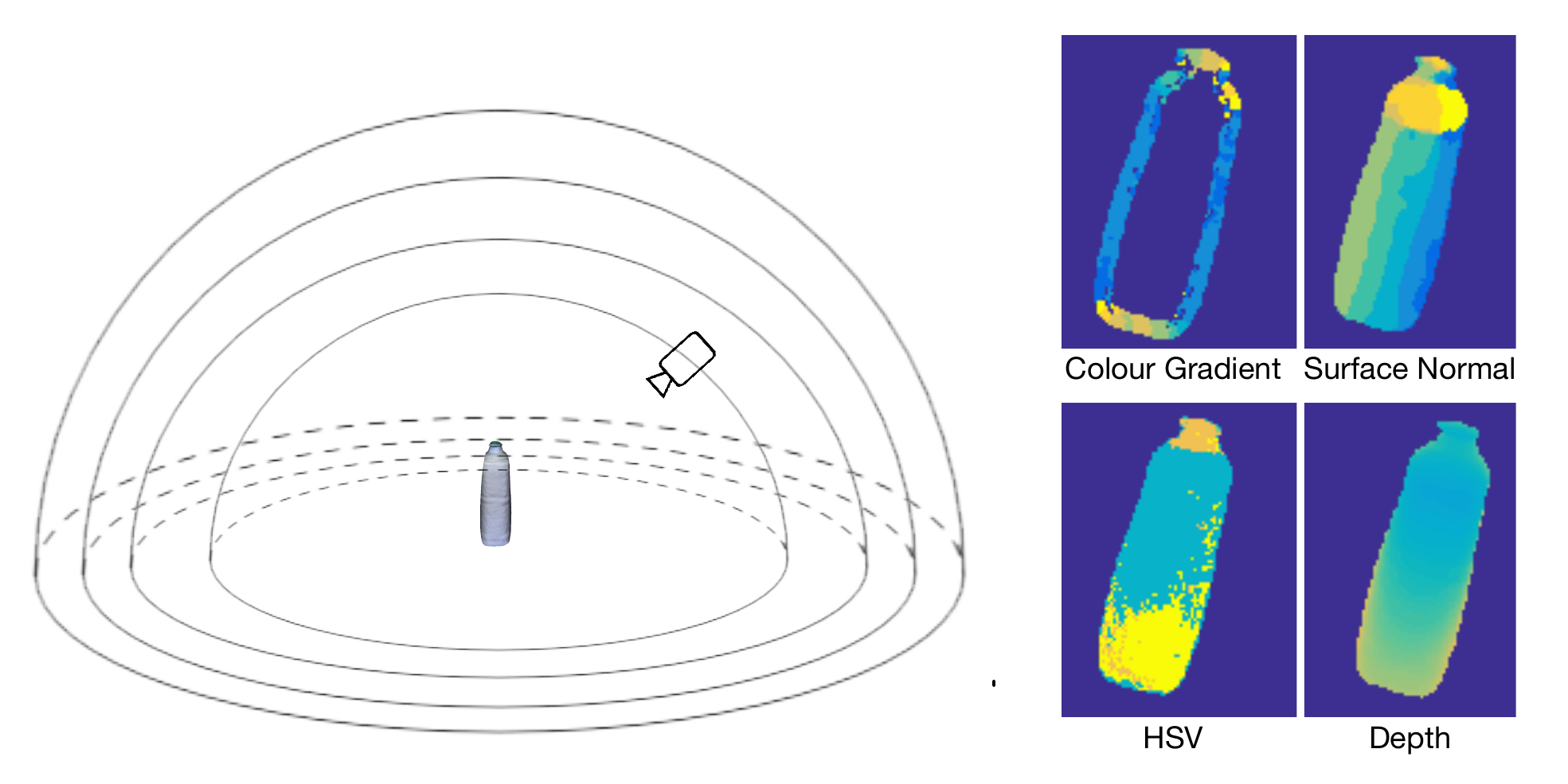}
  \caption{We generate synthetic dataset from object models that provided by LineMOD dataset. The left figure shows a sample procedure of rendering templates on a hemisphere of several radii. Here we use four modalities in this work, from left-top: colour gradient, surface normal, hue colour and depth.}
\label{fig:training}
\end{figure}

\noindent\textbf{Training} Similar to most of recent approaches on 6-dof textureless object pose estimation problem, we synthetically generate our template dataset from 3D object CAD models. The templates are computed from rendered object views on upper hemispheres of several radii, same as the sampling strategy in \cite{Linemod_ACCV}, as shown in Figure~\ref{fig:training}. Each template $\template$ is assigned with a 2-tuple label $(\object,\pose)$, which denotes the object class and object pose (yaw, roll and pitch angles) respectively.

We first expand the template $\template$ into a $| \locationList |$-dimensional descriptor: $\descriptor_{\template} = \{\feat(\refImage,\location):r \in \locationList\}$. In practice, we use LineMOD as our descriptor including an additional object hue map as described in \cite{Linemod_ACCV}:
\begin{equation}
	\feat(\refImage,\location) = \{\text{CG}(\refImage,\location),\text{SN}(\refImage,\location),\text{Hue}(\refImage,\location)\},
\end{equation}
where CG, SN and Hue represent three modalities used in LineMOD: colour gradient, surface normal and hue colour respectively. Each template is therefore a long vector of integers: $\descriptor = (\featValue_{(1,1)}, ..., \featValue_{(| \locationList |,| \modalityList|)} )$, where $\modalityList$ is a list of modality used and $\featValue$ is quantised feature value from 0 to 8. The value 0 appears when the feature is not significant, \eg image colour gradient that norm is below a certain threshold. See \cite{Linemod_ACCV} for the detail of each modality. For the feature out of the object mask, we use uniform noise to model the background. Different background models are evaluated in \cite{RF}.

\begin{figure*}[t!]
\centering
  \includegraphics[width=0.8\textwidth]{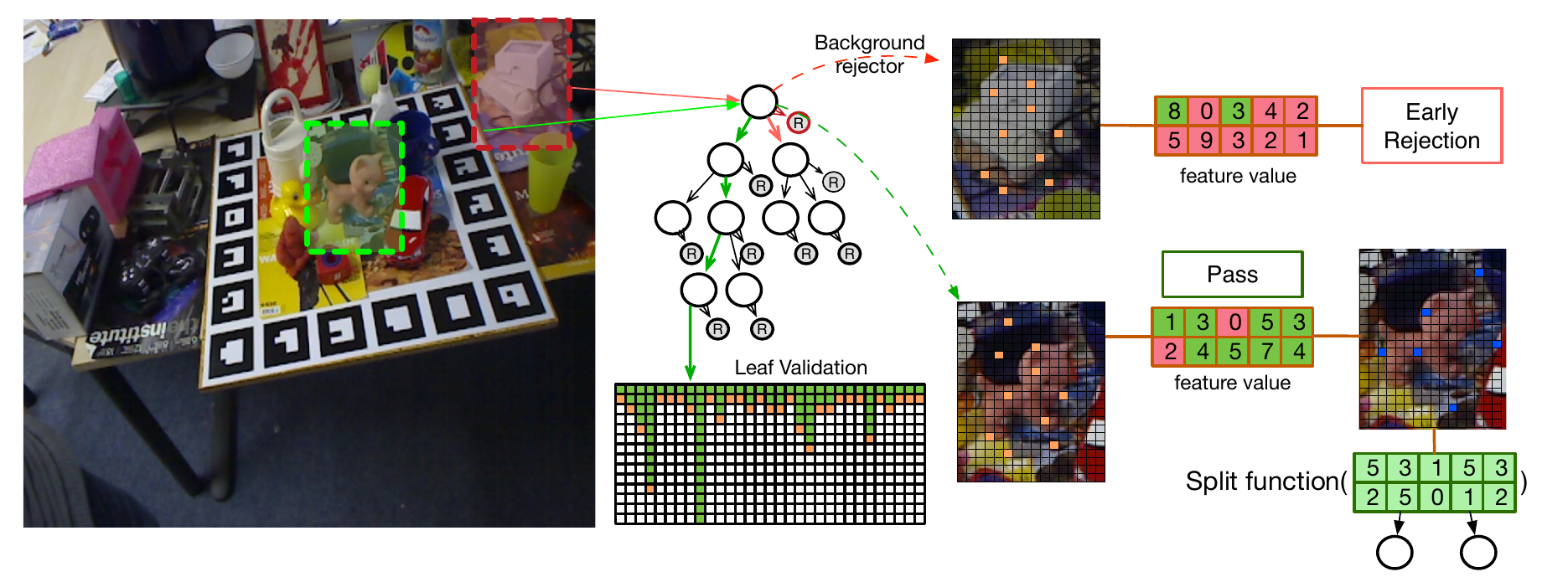}
  \caption{Visualisation of our proposed pipeline. At each candidate sliding window we extract LineMOD feature descriptor and pass to the decision tree. At each node, the extracted features are examined by a preemptive background rejector, the candidate enters full validation stage only if it passes all the rejectors. This process significantly save the time cost on background noises. Finally, we further accelerate the validation stage with a fast breadth-first scheme, inspired by \cite{PRansac}.}
\label{fig:overview}
\end{figure*}
\noindent\textbf{Split Function} A template descriptor set $\templateSet_\nodeIndex$, arriving at $\nodeIndex$th node, is partitioned into two subsets $\templateSet_\nodeIndex^L,\templateSet_\nodeIndex^R$ by a split function $\splitfunc(\descriptor,\splitParam_\nodeIndex) \in \{0,1\}$:
\begin{align} \label{eq:sf_childnodes}
\left\{
                \begin{array}{ll}
                  \templateSet_\nodeIndex^L(\templateSet_\nodeIndex,\splitParam) &= \{\descriptor \in \templateSet_\nodeIndex | \splitfunc(\descriptor,\splitParam) = 0 \}\\
                  \templateSet_\nodeIndex^R(\templateSet_\nodeIndex,\splitParam) &= \{\descriptor \in \templateSet_\nodeIndex | \splitfunc(\descriptor,\splitParam) = 1 \}\\
                \end{array}
\right.
\end{align}

where $\splitParam$ denotes split node parameters. The split node parameter can be denoted as $\splitParam = (\featSelect,\splitPlane,\splitThresh)$ where $\featSelect$ selects a small subspace of entire feature space as feature selector function, $\splitPlane$ defines the geometric primitive used to separate the data, $\splitThresh$ denotes thresholds in the binary test. The parameter is chosen to maximise an energy function, usually the information gain to ensure an optimal split. In practice, the design of split function is crucial to achieve good performance. In the later section we will discuss the impact of different split function on template matching performance.

\noindent\textbf{Leaf Validation}
The training templates are recursively split until meets stopping criteria. This involves the control of tree shape, depth and thus the trade-off between the generalisation power and efficiency. As briefly explained in previous section, full validation on a template set is expensive and almost always the bottleneck of the whole pipeline as in many recent related works. Ideally we want to keep the number of templates in leaf nodes as few as possible while avoiding the overfitting. 

In practice when we apply tree-based search on standard template matching, no matter how we set the stopping criteria, the lack of training data and high dimensional feature is likely to lead to overfitting. One possible workaround is a variant of \kdtree approach (Best-bin-first) \cite{kdtree_bbf}, which backtracks from the leaf node according to a priority queue based on the closeness between query and the bin boundary, until a fixed number of nearest candidates is searched. However, this method is less efficient when large outlier presents as the closeness is no longer reliable. Also, the optimal number of nearest candidates from backtracking varies with object class and can only be decided empirically. We will also address this issue later in our method.

\subsection{Split Function for Insufficient, High-dimensional Noisy Data}
In many applications, feature selector $\featSelect(\descriptor) \in \mathbb{R}^{\dimension'}$, where often $\dimension' = 1$ or $2$ is sufficient. However, more dimensions are needed to compensate for the less distinctive, heavily quantised features and higher outlier ratio. 

Figure~\ref{fig:overview} shows an overview of our method. To avoid the superlinear time cost from randomised node optimisation due to high dimensionality, we randomly draw an exemplar $\exemplar$ from the template set $\templateSet$, such that:
\begin{align} \label{eq:sf_thresh}
	\splitfunc(\descriptor,\splitParam) &= \left\{
                \begin{array}{ll}
                  1, &  \text{if } (\| \featSelect(\exemplar) - \featSelect(\descriptor)\| < \splitThresh)\\
                  0, & \text{otherwise}\\
                \end{array}
              \right.\\
	\nonumber\featSelect(\descriptor) &= (\featValue_{\featSelect_1},\featValue_{\featSelect_2},...,\featValue_{\featSelect_{\dimension'}}), \featSelect_i \in [1,\dimension]
\end{align}
which maximises the energy function $\energyFunc$:
\begin{align}
	\splitParam_\nodeIndex = \argmax_{(\splitParam \in \splitParamPool)} \energyFunc(\templateSet_\nodeIndex,\splitParam),
\end{align}
where $\energyFunc$ denotes the energy function, $\splitParamPool$ denotes a randomly generated set from the entire parameter space. Since the space is greatly reduced with the exemplar approach, the size of $\splitParamPool$ should be limited to a small number to maintain the efficiency.

For the choice of energy function, we modify the standard unsupervised entropy to cope with the missing feature values:

\begin{align}
	\energyFunc(\templateSet,\splitParam) &= \sum_{\descriptor \in \templateSet}\completeness(\featSelect(\descriptor)) * \informationGain(\templateSet,\splitParam)\\
	\nonumber\completeness(\descriptor) &=  \sum_{\featValue \in \descriptor}{\objMask(\featValue) \in \{0,1\}}\\
	\nonumber\informationGain(\templateSet,\splitParam) &= \entropy(\templateSet,\splitParam) - \frac{|\templateSet^L| \entropy(\templateSet^L,\splitParam) + |\templateSet^R| \entropy(\templateSet^R,\splitParam)}{|\templateSet|}
\end{align}
where $\objMask$ denotes a foreground boolean indicator such that $\objMask(\featValue) = 1$ if $\featValue$ located on the object mask on the template and vice versa; $\entropy$ denotes an entropy function. In template matching or NN problem in general, each data point is assigned to a unique label. Therefore the standard entropy function is not suitable here. Instead we use an unsupervised variant:
\begin{align}
	\entropy(\templateSet,\splitParam) &= -\sum_{i \in \featSpace} \featCount(\templateSet,i,\splitParam) \log \featCount(\templateSet,i,\splitParam) \\
	\nonumber\featCount(\templateSet,i,\splitParam) &= \frac{1}{|\templateSet|  |\featSelect(\descriptor)|}\sum_{\descriptor \in \templateSet}\sum_{\featValue \in \featSelect(\descriptor)}\isEqual(\featValue,i) \\
	\nonumber\isEqual(\featValue,i) &= \left\{
                \begin{array}{ll}
                  1, &  \text{if } (\featValue = i)\\
                  0, & \text{otherwise}\\
                \end{array}
              \right.
\end{align}
where $\featSpace$ is the feature space, \ie $\featSpace$ = \{1,2,...8\} in the case of LINE-MOD descriptor. The entropy measures the uncertainty
associated with the feature values given the feature dimension, a higher entropy yields better separation.

Next, we adapt a simple fuzzy rule to on the thresholding to deal with insufficient data. This approach has been proposed in the literature \cite{fuzzy_another,fuzzy_origin} but not drawn much attention from the field of computer vision. This adaptation tends to tolerate imprecise, missing feature values and reduce the classification ambiguity from the split function, achieved by duplication of feature vectors that near to the split subspace. 

We modify the binary test in equation \ref{eq:sf_childnodes} and \ref{eq:sf_thresh} such that:
\begin{align} \label{eq:sf_modified}
&\left\{
                \begin{array}{ll}
                  \templateSet_\nodeIndex^L(\templateSet_\nodeIndex,\splitParam)&= \{\descriptor \in \templateSet_\nodeIndex | \splitfunc(\descriptor,\splitParam) < \fuzzyFactor \}\\
                  \templateSet_\nodeIndex^R(\templateSet_\nodeIndex,\splitParam)&= \{\descriptor \in \templateSet_\nodeIndex | \splitfunc(\descriptor,\splitParam) > - \fuzzyFactor \}\\
                \end{array}
\right.\\
&\splitfunc(\descriptor,\splitParam) = \| \featSelect(\exemplar) - \featSelect(\descriptor)\| - \splitThresh, 
\end{align}

thus feature vectors that fall into the `fuzzy' interval $[- \fuzzyFactor,\fuzzyFactor]$ will be passed to both child nodes. This approach allows feature vectors to reach multiple leaves, which greatly reduces the overfitting due to lack of training data.

\subsection{Preemptive Background Rejector}
A fast coarse estimation of objectness is common in many detection methods, since the object of interest generally occupies only a small portion of the testing image. We further propose a preemptive background rejector as an extra split function in each node that sends the query to a `background' leaf node if it fails a binary test. In contrast to most of the background removal methods, our approach does not exploit negative samples. Instead, we make assumption that for all feature vectors that do not exist in the dataset are negative samples. Here we isolate the foreground from background by minimising the entropy in the rejector function so that all foreground feature vectors share similar values:

\begin{align}
	\rsplitfunc(\descriptor,\rsplitParam) &= \left\{
                \begin{array}{ll}
                  1, &  \text{if } (\frac{1}{|\featSelect(\descriptor)|}\sum_{\featValue \in \rfeatSelect(\descriptor)} \isBackground(\featValue,\rsplitParam) < \rsplitThresh)\\
                  0, & \text{otherwise}\\
                \end{array}
              \right.\\
	\rsplitParam_\nodeIndex &= \argmin_{(\rsplitParam_\nodeIndex \in \splitParamPool)} \sum_{\descriptor \in \templateSet} \informationGain(\templateSet,\splitParam),
\end{align}
which $\rsplitThresh$ denotes a threshold in $[0,1]$ to control the acceptance of outlier ratio; $\isBackground$ denotes a background feature look-up table, such that:
\begin{align}
	\isBackground(\templateSet_\nodeIndex,i,\rsplitParam) = \left\{
                \begin{array}{ll}
                  1, &  \text{if } \featCount(\templateSet_\nodeIndex,i,\rsplitParam) >  \rLUTThresh\\
                  0, & \text{otherwise}\\
                \end{array}
              \right.
\end{align}

The query $\descriptor$ is rejected immediately at a node $\nodeIndex$ if $\rsplitfunc(\descriptor,\rsplitParam_\nodeIndex) = 1$.

\subsection{Fast Breadth-first Leaf Validation}
The leaf nodes contain only tens or at most a hundred templates, however pairwise matching all candidates is still computational expensive. In practice, most of bad candidates can be safely removed by examining only a small portion from the whole feature descriptor. Therefore we propose a further speedup of the validation process with a breadth-first preemption scheme inspired by preemptive RANSAC \cite{PRansac}. 

\begin{figure}[h]
\centering
  \includegraphics[width=0.5\textwidth]{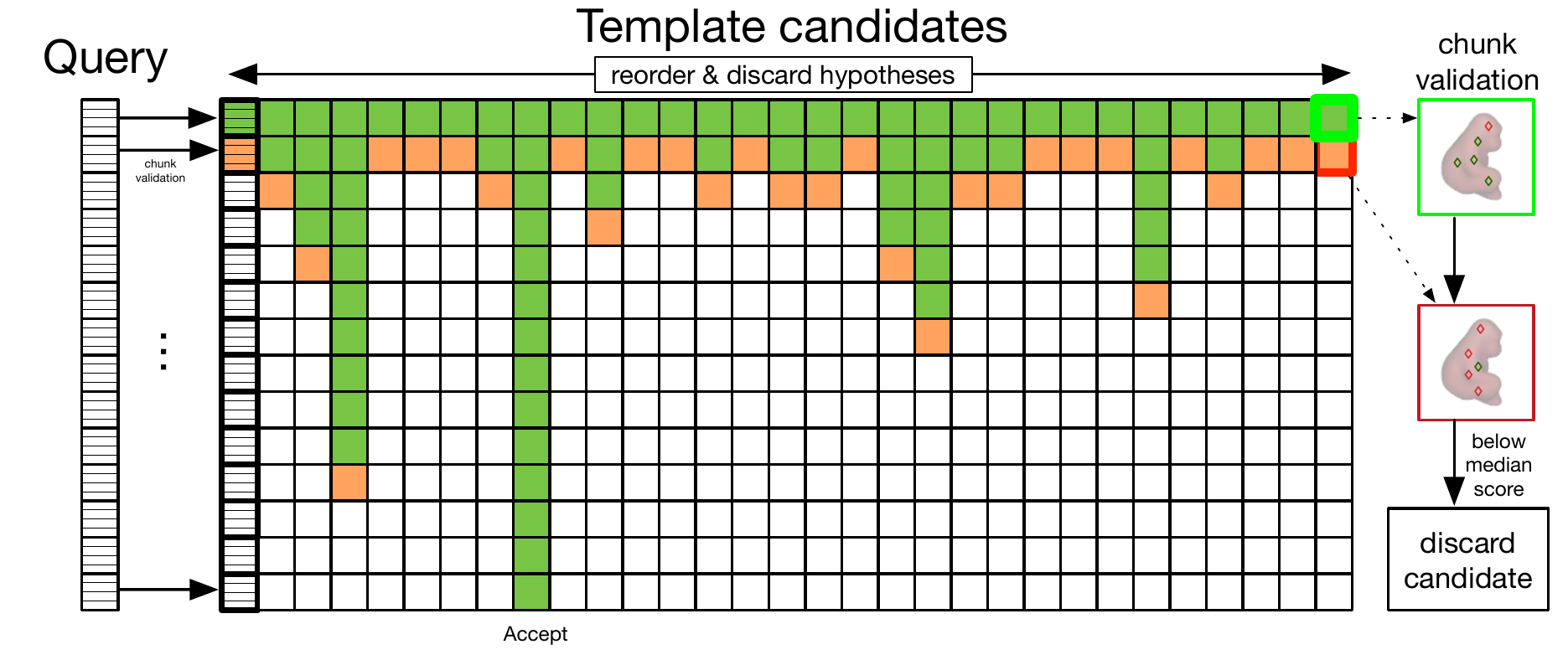}
  \caption{Breath-first preemptive scheme for leaf validation speed up. The templates are equally split into small chunks to alleviate the time cost from validating bad candidates.}
\label{fig:pransac}
\end{figure}

As shown in Figure~\ref{fig:pransac}, during leaf validation, we equally split template feature descriptors into smaller chunks, each contains a fixed number of features. In each stage, we score and compare all chunks and keep only the candidates that satisfy the pass condition. Scores are accumulated to the next stage and repeat until there is only one or no candidate left.

In our real-time implementation, we use the pass condition

\begin{align}
	\function(\descriptor') &= \left\{
                \begin{array}{ll}
                  1, &  \text{if } (\scoring(\descriptor') > \max(\text{Median}(\scoring(\descriptor)),\scoringConst)\\
                  0, & \text{otherwise},
                \end{array}
              \right.
\end{align}

where \scoring(\descriptor) is a scoring function that measures the distance between query and candidate, $\scoringConst$ is a constant threshold. We set the threshold $\scoringConst = 0.5$ empirically. With this approach, the validation time complexity is reduced to $\bigo(\log n)$. The chunk size works as a trade-off between accuracy and speed: larger chunk leads to a better robustness but less efficiency, and vice versa. Furthermore, we add depth value as another modality in the validation stage to measure shape similarity.

\begin{table*}[!ht]
\centering
\begin{tabular}{l|llll|l|llll}
\hline
1 Tree  & T\_Tree & T\_Valid & T\_Total & Acc.   & 5 Trees & T\_Tree & T\_Valid & T\_Total & Acc.   \\ \hline
ape     & 0.20 ms & 6.50 ms  & 6.70 ms  & 96.0\% &         & 0.99 ms & 12.31 ms & 13.30 ms & 97.1\% \\
bvise   & 0.43 ms & 13.37 ms & 13.80 ms & 91.1\% &         & 2.13 ms & 29.50 ms & 31.63 ms & 93.2\% \\
cam     & 0.41 ms & 11.70 ms & 12.11 ms & 93.1\% &         & 1.93 ms & 24.66 ms & 26.59 ms & 94.8\% \\
can     & 0.44 ms & 13.07 ms & 13.51 ms & 91.5\% &         & 2.22 ms & 28.45 ms & 30.67 ms & 92.0\% \\
cat     & 0.23 ms & 8.34 ms  & 8.57 ms  & 94.3\% &         & 1.05 ms & 15.23 ms & 16.28 ms & 95.5\% \\
driller & 0.39 ms & 13.93 ms & 14.32 ms & 95.4\% &         & 1.73 ms & 29.01 ms & 30.74 ms & 96.0\% \\
duck    & 0.28 ms & 9.64 ms  & 9.92 ms  & 90.0\% &         & 1.20 ms & 15.99 ms & 17.19 ms & 94.5\% \\
eggbox  & 0.30 ms & 10.73 ms & 11.03 ms & 98.3\% &         & 1.18 ms & 20.13 ms & 21.31 ms & 98.9\% \\
glue    & 0.34 ms & 11.46 ms & 11.80 ms & 92.1\% &         & 1.62 ms & 29.85 ms & 31.47 ms & 94.4\% \\
hpunch  & 0.44 ms & 14.76 ms & 15.20 ms & 90.7\% &         & 2.19 ms & 33.12 ms & 35.31 ms & 93.6\% \\
iron    & 0.39 ms & 11.82 ms & 12.21 ms & 91.9\% &         & 1.67 ms & 19.38 ms & 21.05 ms & 92.7\% \\
phone   & 0.40 ms & 13.93 ms & 14.33 ms & 89.8\% &         & 1.97 ms & 29.44 ms & 31.41 ms & 91.0\% \\ \hline
Average & 0.35 ms & 11.60 ms & 11.96 ms & 92.9\% &         & 1.66 ms & 23.92 ms & 25.58 ms & 94.4\% \\ \hline\hline
Hashmod\cite{Hashmod} & - & - & 83 ms & 96.5\% & DTT-3D\cite{DTT}  & - & - & 55 ms & 97.2\% \\ \hline
LineMOD\cite{Linemod_ACCV} & - & - & 119 ms & 96.6\% &  &  &  & & \\ \hline
\end{tabular}

\caption{Accuracy and average time per frame for the whole pipeline with a 1 \& 5 trees. Our approach is several times faster than the state-of-the-art approaches with comparable accuracy. } 
\label{table:time_1}
\end{table*}

\begin{table}[!ht]
\centering
\begin{tabular}{lllll}
                              &  5 Objects        &                             &  15 Objects        &        \\
\multicolumn{1}{l|}{1 Object} & T\_Total & \multicolumn{1}{l|}{Acc.}   & T\_Total & Acc.   \\ \hline
\multicolumn{1}{l|}{ape}      & 20.01 ms & \multicolumn{1}{l|}{97.3\%} & 25.53 ms & 97.4\% \\
\multicolumn{1}{l|}{bvise}    & 50.30 ms & \multicolumn{1}{l|}{94.4\%} & 60.11 ms & 93.4\% \\
\multicolumn{1}{l|}{cam}      & 59.81 ms & \multicolumn{1}{l|}{94.7\%} & 64.47 ms & 94.0\% \\
\multicolumn{1}{l|}{can}      & 53.54 ms & \multicolumn{1}{l|}{93.3\%} & 66.98 ms & 93.1\% \\
\multicolumn{1}{l|}{cat}      & 34.09 ms & \multicolumn{1}{l|}{95.2\%} & 45.22 ms & 96.0\% \\
\multicolumn{1}{l|}{driller}  & 66.40 ms & \multicolumn{1}{l|}{96.4\%} & 78.94 ms & 95.8\% \\
\multicolumn{1}{l|}{duck}     & 29.19 ms & \multicolumn{1}{l|}{95.5\%} & 42.23 ms & 96.2\% \\
\multicolumn{1}{l|}{eggbox}   & 34.50 ms & \multicolumn{1}{l|}{98.8\%} & 47.98 ms & 98.6\% \\
\multicolumn{1}{l|}{glue}     & 52.19 ms & \multicolumn{1}{l|}{94.7\%} & 65.99 ms & 94.6\% \\
\multicolumn{1}{l|}{hpunch}   & 56.32 ms & \multicolumn{1}{l|}{95.2\%} & 74.56 ms & 95.3\% \\
\multicolumn{1}{l|}{iron}     & 32.13 ms & \multicolumn{1}{l|}{93.2\%} & 44.72 ms & 93.6\% \\
\multicolumn{1}{l|}{phone}    & 54.71 ms & \multicolumn{1}{l|}{93.3\%} & 56.10 ms & 93.3\% \\ \hline
\multicolumn{1}{l|}{Average}  & 45.27 ms & \multicolumn{1}{l|}{95.2\%} & 56.07 ms & 95.1\% \\ \hline\hline
\multicolumn{1}{l|}{Hashmod\cite{Hashmod}}  & 131 ms &  \multicolumn{1}{l|}{95.5\%} & 184 ms & 95.1\% \\
\multicolumn{1}{l|}{DTT-3D\cite{DTT}} & 107 ms &  \multicolumn{1}{l|}{97.2\%} & 239 ms & 97.2\% \\
\multicolumn{1}{l|}{LineMOD\cite{Linemod_ACCV}}  & 427 ms &  \multicolumn{1}{l|}{96.6\%} & 1197 ms & 96.6\% \\ \hline
\end{tabular}
\caption{Accuracy and average time per frame for multiple object (5 trees) and its comparison with state-of-the-art approaches.}
\label{table:time_2}
\end{table}

\subsection{Evaluation}

Experiments are conducted on LineMOD ACCV12 dataset \cite{Linemod_ACCV} consisting of 13 object CAD models and testing images for object detection and 6D pose estimation. We apply the same evaluation criteria with a distance factor of $k_m = 0.1$. We run the experiments on a single 2.8 GHz Intel Core i7, the whole forest takes around 10 MB. A pyramid scheme is applied in a similar way to \cite{Linemod_ACCV}.

In overall, our pipeline achieve sublinear time complexity and comparable high accuracies as shown in Table~\ref{table:time_1}. Despite the detect rate of our approach is marginally worse than state-of-the-arts, we are at least two times faster than the fastest DTT-3D \cite{DTT}. Since our approach is tree-based, it is also scalable to more objects. Table~\ref{table:time_2} shows that we significantly outperform state-of-the-art approaches in speed with more objects. Additionally, both table show that our approach works favourably on simpler object, because the training templates share more similar features so the background is more likely to be rejected before the expensive validation stage. 

\begin{figure}[h]
\centering
  \includegraphics[width=0.5\textwidth]{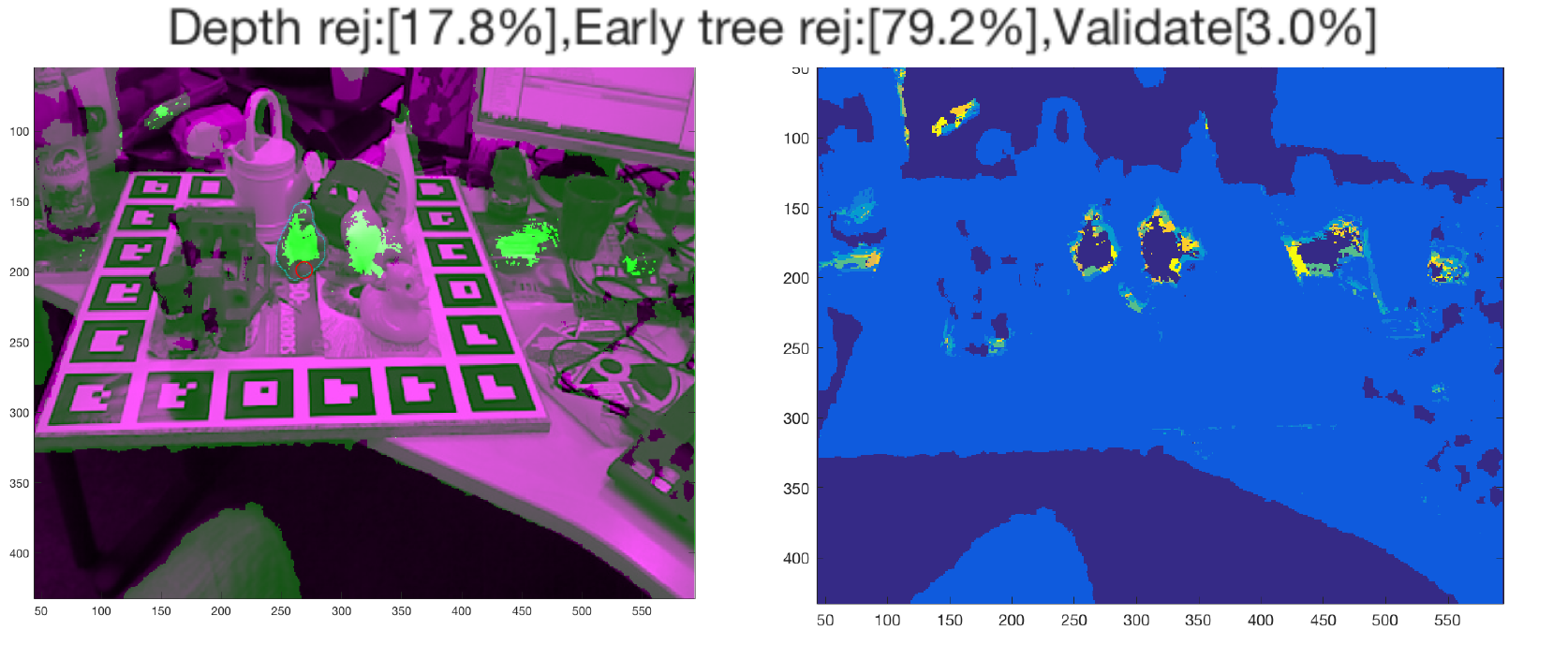}
  \caption{in this sample frame, with our proposed preemptive background rejector up to 97\% negative bounding boxes are filtered before reaching the forest leaf nodes. The green region in left image indicates the locations that enter the validation stage; the right image shows the tree depth when background locations are rejected, dark blue indicates the region is either rejected due to out-of-depth-range or pass the validation stage, light blue to yellow indicates whether the locations are rejected early or late in the forest. As nearer to the ground truth or ambiguous objects the location is more likely to not be rejected earlier.}
\label{fig:testingER}
\end{figure}

In Figure~\ref{fig:testingER}, we illustrate the effectiveness of our proposed preemptive background rejector. Depending on the object complexity and scene, up to 90\% to 97\% background locations can be filtered out before validation stage with a very high recall rate. Since textureless objects are generally simple in shape and colour, their rendered templates are likely to share particularly features that can easily rule out background clutters. 

Since most candidate locations are rejected at the first node in the decision tree, from Table~\ref{table:time_1} we can see that the time cost on testing tree itself (T\_Tree) is negligible compare to the validation time (T\_Valid). The performance is further boosted with multiple trees with random permutation. Here our proposed approach shows another advantage with random forest framework. The time cost for additional trees has sublinear growth as the leaf nodes in each tree are concatenated to remove duplications before entering the validation stage. The result shows that the accuracy is increased by 1.5\% with 5 trees but only approximately 2 times slower.

For the adaption of fast breadth-first leaf validation, the time taken is reduced up to 4 or 5 times without loss of accuracy. Note that in practice, due to the overhead this approach does not works well if the set of template is too small, in our case, we set our maximum tree depth to be 8, and 9 for multiple objects.

Finally, our approach achieves sublinear complexity as the result of tree structure and significantly outperform all state-of-the-art approaches in speed.

\subsection{Conclusion}
We present an efficient and scalable approach for 3D object detection and pose estimation which modifies the randomised forest framework to cope with background noises. The result shows that we significantly outperform the state-of-the-art methods in terms of speed while maintaining a reasonable recognition accuracy. This approach can be generalised to any machine learning task that has very low positive rate, such as for typical face (or object) detection problem, on average only 0.01\% of all sub-windows are positive. This assumption is especially true in most real-world applications.

Moreover, this approach is limited only to random forest framework, it has potential to be implemented to any directed-acyclic-graph-(DAG)-based classifier, such as deep convolutional neural network (CNN). It is widely known that typical CNN forward propagation consumes very high computational power and require specialised hardware (GPUs) with high power consumption, often around 250W per card. This motivates the implementation of early termination operator to CNN framework.
{\small
\bibliographystyle{ieee}
\bibliography{egbib}
}

\end{document}